%% file: main-5043-Wang.tex
\documentclass[11pt,a4paper]{article}
\usepackage{times,latexsym}
\usepackage{url}
\usepackage{xspace}
\usepackage{paralist}
\usepackage[T1]{fontenc}

\usepackage[acceptedWithA]{tacl2021v1}

\usepackage{xspace,mfirstuc,tabulary}

\newcommand{\ex}[1]{\textit{#1}\xspace}
\newcommand{\z}{\phantom{0}}

\newif\iftaclinstructions
\taclinstructionsfalse 
\iftaclinstructions
\renewcommand{\confidential}{}
\renewcommand{\anonsubtext}{(No author info supplied here, for consistency with
TACL-submission anonymization requirements)}
\newcommand{\instr}
\fi

\iftaclpubformat 

\else

\fi

\usepackage{import}
\usepackage{microtype}
\usepackage{layout}
\usepackage{tabularx, makecell}
\usepackage{booktabs}
\usepackage{mathrsfs}
\usepackage{amssymb} 
\usepackage{url}
\usepackage{graphicx}
\usepackage{xspace,paralist}
\usepackage{times,latexsym}
\usepackage{amsmath}
\usepackage{appendix}
\usepackage{comment} 
\usepackage{enumitem}
\usepackage{makecell}
\usepackage{multirow}
\usepackage{xcolor}
\usepackage{graphicx}
\usepackage{cleveref}

\newcommand{\eqnref}[1]{Eq~\eqref{#1}\xspace}
\newcommand{\tabref}[2][]{Table#1~\ref{#2}\xspace}
\newcommand{\figref}[1]{Figure~\ref{#1}\xspace}
\newcommand{\secref}[1]{Section~\ref{#1}\xspace}

\newcommand{\metric}[1]{\textsc{#1}\xspace}
\newcommand{\ece}{\metric{ECE}}
\newcommand{\nlpd}{\metric{NLPD}}
\newcommand{\kl}{\metric{KL}}
\newcommand{\NA}{\textsc{n/a}}

\newcommand{\dataset}[1]{\text{#1}\xspace}
\newcommand{\sickr}{\dataset{SICK-R}}
\newcommand{\unli}{\dataset{UNLI}}
\newcommand{\stsb}{\dataset{STS-B}}

\newcommand{\tedx}{\dataset{TED-X}}
\newcommand{\xnli}{\dataset{XNLI}}
\newcommand{\paws}{\dataset{PAWS}}
\newcommand{\pawsx}{\dataset{PAWS-X}}
\newcommand{\usts}{\dataset{USTS}}
\newcommand{\ustsu}{\dataset{USTS-U}}
\newcommand{\ustsc}{\dataset{USTS-C}}

\newcommand{\model}[1]{\text{#1}\xspace}
\newcommand{\mcdropout}{\model{MC-Dropout}}
\newcommand{\simcse}{\model{SimCSE}}

\title{Collective Human Opinions in Semantic Textual Similarity}

\author{
Yuxia Wang$^\spadesuit$ 
   \qquad Shimin Tao$^\clubsuit$ 
   \qquad Ning Xie$^\clubsuit$ 
   \qquad Hao Yang$^\clubsuit$ \\
   \qquad Timothy Baldwin$^{\spadesuit\heartsuit}$ 
   \qquad Karin Verspoor$^{\spadesuit\diamondsuit}$
  \\
  $\spadesuit$  
  The University of Melbourne, Melbourne, Victoria, Australia \\
  $\clubsuit$ Huawei TSC, Beijing, China \qquad
  $\heartsuit$  
  MBZUAI, Abu Dhabi, UAE \\
  $\diamondsuit$RMIT University, Melbourne, Victoria, Australia \\
  \texttt{yuxiaw@student.unimelb.edu.au \,\,\, karin.verspoor@rmit.edu.au} \\
  \texttt{tb@ldwin.net \,\,\, \{taoshimin,nicolas.xie,yanghao30\}@huawei.com} \\
  \texttt{}
}

\date{}

\let\oldmaketitle\maketitle
\renewcommand{\maketitle}{\oldmaketitle\setcounter{footnote}{-1}}

\begin{document}
\maketitle
\begin{abstract}
Despite the subjective nature of semantic textual similarity (STS) and pervasive disagreements in STS annotation, existing benchmarks have used averaged human ratings as gold standard.
Averaging masks the true distribution of human opinions on examples of low agreement, and prevents models from capturing the semantic vagueness that the individual ratings represent.
In this work, we introduce \usts, the first \textbf{U}ncertainty-aware \textbf{STS} dataset with $\sim$15,000 Chinese sentence pairs and 150,000 labels, to study collective human opinions in STS.
Analysis reveals that neither a scalar nor a single Gaussian fits a set of observed judgements adequately.
We further show that current STS models cannot capture the variance caused by human disagreement on individual instances, but rather reflect the predictive confidence over the aggregate dataset.
\end{abstract}

\input{section/intro}
\input{section/background}
\input{section/datacollection}
\input{section/humanjudgementanalysis}
\input{section/prediction}
\input{section/multilingual}
\input{section/discussion}
\input{section/conclusion}
\input{section/acknowledgement}

\bibliography{tacl2021}
\bibliographystyle{acl_natbib}

\end{document}

%% file: section/intro.tex
\section{Introduction}
\label{sec:introduction}
Semantic textual similarity (STS) is a fundamental natural language understanding (NLU) task, involving the prediction of the degree of semantic equivalence between two pieces of text (S1,S2).
STS has been approached in various ways, ranging from early efforts using string- or knowledge-based measures and count-based co-occurrence models~\citep{resnik1999semantic, DBLP:conf/coling/Barron-CedenoRAL10, matveeva2005generalized}, to modern neural networks. 

Broadly speaking, the goal of the STS task is to train models to make a similarity assessment that matches what a human would make. Gold-standard scores are typically assigned by asking multiple raters to label a pair of sentences and then taking the average~\citep{agirre2012semeval, agirre2013sem, agirre2014semeval, agirre2015semeval, agirre2016semeval, marelli2014semeval, souganciouglu2017biosses, wang2018medsts}. 
The underlying assumption here is that there is a single ``true'' similarity score between S1 and S2, and that this label can be approximated by averaging multiple --- possibly noisy --- human ratings. 
While this assumption might be reasonable in settings such as educational testing with well-defined knowledge or norms~\citep{trask1999key}, it is not the case for more subjective NLU tasks. 

\citet{pavlick-kwiatkowski-2019-inherent} show that in natural language inference (NLI), disagreements often persist even if more ratings are collected or when the amount of context provided to raters is increased.
High disagreement has been observed in a number of existing NLI datasets~\cite{nie-etal-2020-learn}.
In STS, concerns about inconsistent judgements have been raised, particularly for difficult boundary cases in complex domains, where even expert annotators can disagree about the ``true'' label~\citep{wang2020n2c2sts, olmin2022robustness}.
Identifying and discarding ``noisy'' labels during training can reduce generalisation error~\citep{wang2022noisy, wang2022uncertaintysts}. 
We reexamine whether the disagreement observed among raters should be attributed to ``noise'' and resolved via dismissing, or should rather treated as an inherent quality of the STS labels.
Specifically, our primary contributions are: 
\begin{compactenum}
    \item We develop \usts, the first \textbf{U}ncertainty-aware \textbf{STS} dataset with a total of $\sim$15,000 Chinese sentence pairs and 150,000 labels. We study the human assessments and investigate how best to integrate them into a gold label across varying degrees of observed human disagreement.
    \item We show that state-of-the-art STS models cannot capture disagreement when trained using a single averaged rating, and argue that STS evaluation should incentivise models to predict distributions over human judgements, especially for cases of low agreement.    
    \item We discuss the practicalities of transferring labels across languages in building a multilingual STS corpus, and present evidence to suggest that this may be problematic in the continuous labelling space. 
\end{compactenum}

%% file: section/background.tex
\section{Background}
\label{sec:background}
\subsection{Semantic Textual Similarity Task}
\label{sec:ststask}

\textbf{Data Collection and Annotation:}
As STS requires a sentence pair, to construct a dataset, ideally sentence pairs should be sampled to populate the spectrum of differing degrees of semantic equivalence, which is a huge challenge. If pairs of sentences are taken at random, the vast majority would be totally unrelated, and only a very small fraction would have some degree of semantic equivalence~\citep{agirre2012semeval}. Accordingly, previous work has either resorted to string similarity metrics (e.g.\ edit distance or bag-of-word overlap) \citep{agirre2013sem, agirre2014semeval, agirre2015semeval, agirre2016semeval, souganciouglu2017biosses, wang2018medsts}, or reused existing datasets from tasks related to STS, such as paraphrasing based on news/video descriptions~\citep{agirre2012semeval} and NLI~\citep{marelli2014semeval}.

In terms of annotation, for general text (e.g.\ news, glosses, or image descriptions), it has mostly been performed using crowdsourcing via platforms such as Amazon Mechanical Turk with five crowd workers \citep{cer2017semeval}. For knowledge-rich domains such as clinical and biomedical text, on the other hand, a smaller number of expert annotators has been used, such as two clinical experts for MedSTS~\citep{wang2018medsts}. 
Raters are asked to assess similarity independently on the basis of semantic equivalence using a continuous value in range $[0,5]$. 
Then a gold label is computed by averaging these human ratings.

\begin{table}[!t]
\centering
\resizebox{\columnwidth}{!}{
    \begin{tabular}{l p{7cm}}
        \toprule
        \textbf{No.\ 1} & \textsc{Low Human Disagreement} \\[1ex]
        S1 & \ex{Kenya \underline{Supreme Court} upholds election result.} \\
        S2 & \ex{Kenya \underline{SC} upholds election result.} \\
        Old label & 5.0 \\
        New label & $\mathcal{N}(\mu=4.9,\sigma=0.1)$ \\[1ex]
        Annotations & [4.5, 4.7, 4.8, 5.0, 5.0, 
                       5.0, 5.0, 5.0, 5.0, 5.0,
                       5.0, 5.0, 5.0, 5.0, 5.0] \\
        Prediction & 3.5 \\
        Reason & Lack of knowledge of the correspondence between \textit{Supreme Court} and \textit{SC}. \\
        \midrule
        \textbf{No.\ 2} & \textsc{High Human Disagreement} \\[1ex]
        S1 & \ex{A man is carrying a canoe with a dog.} \\
        S2 & \ex{A dog is carrying a man in a canoe.}  \\
        Old label & 1.8 \\
        New label & $\mathcal{N}(\mu=1.7,\sigma=1.0)$ \\[1ex]
        Annotations & [0.0, 0.3, 0.5, 0.5, 1.2, 
                       1.5, 1.5, 1.8, 2.0, 2.0, 
                       2.0, 2.0, 2.5, 3.5, 3.5] \\
        Prediction & 4.3 \\
        Reason & Uncertainty about the impact of key differences in
                 event participants on instances of high lexical overlap \\
        \midrule
        \textbf{No.\ 3} & \textsc{High Human Disagreement} \\[1ex]
        S1 & \ex{Someone is grating a \underline{carrot}.} \\
        S2 & \ex{A woman is grating an \underline{orange food}.} \\
        Old label & 2.5 \\
        New label & $\mathcal{N}(\mu=2.4,\sigma=1.1)$ \\[1ex]
        Annotations & [0.5, 1.0, 1.0, 1.8, 1.8, 
                       1.8, 2.0, 2.2, 2.5, 3.0,
                       3.0, 3.2, 3.5, 3.6, 4.5] \\
        Prediction & 0.6 \\
        Reason & Failure to associate \textit{carrot} with \textit{orange food}. \\
        \bottomrule
    \end{tabular}``
    }
    \caption{Examples with varying levels of human disagreement from the STS-B validation set. ``Old label'' = gold label of STS-B; ``New label'' = full distribution aggregated by 15 new ratings;
    and ``Prediction'' = similarity score predicted by SOTA STS model.\footnotemark{}}
    \label{tab:examplesofvaryingdisagreements}
\end{table}
\addtocounter{footnote}{-1} 
\stepcounter{footnote}\footnotetext{The individual annotations for STS-B are not available, so we collected new ratings from 15 PhD NLPers. \textit{bert-base} fine-tuned on the STS-B training data ($r$=0.91) is used for prediction, same as the one in \secref{sec:ted} for selection.}

\textbf{Is averaging appropriate?}
Averaging has been the standard approach to generating gold labels since \citet{lee2005empirical}.
However, this approach relies on the assumption that \textit{there is a well-defined gold-standard interpretation + score, and that any variance in independent ratings is arbitrary rather than due to systematic differences in interpretation}.
An example of this effect can be seen in case No.\ 1 in \tabref{tab:examplesofvaryingdisagreements}.
In practice, however, high levels of disagreement can be observed among annotators in different domains.\footnote{$\sigma>0.5$ for 9\% and 11\% pairs in biomedical STS corpora: BIOSSES and EBMSASS; inter-annotator agreement Cohen's $\kappa$=0.60/0.67 for two clinical datasets~\citep{wang2020n2c2sts}.} 

In such cases, a simple average fails to capture the latent distribution of human opinions/interpretations, and masks the uncertain nature of subjective assessments.
With Nos.\ 2 and 3 in \tabref{tab:examplesofvaryingdisagreements}, e.g., the average scores $\mu$ of {1.7} and {2.4} do not convey the fact that the ratings vary substantially ($\sigma > 1.0$).
While the integrated score may reflect the average opinion, it neither captures the majority viewpoint nor exposes the inherent disagreements among raters.
Put differently, \textbf{not all average scores of a given value convey the same information}.
Consider three scenarios that all average to 3.0: (3,3,3,3,3)/5, (1,3.5,3.5,3.5,3.5)/5, and (2,4,2,4,3)/5. The inherent level of human agreement varies greatly in these three cases.

Looking to the system predictions, the model prediction of 3.5 for No.\ 1 in \tabref{tab:examplesofvaryingdisagreements} is clearly incorrect, as it lies well outside the (tight) range of human annotations in the range $[4.5,5.0]$.
While the model prediction of 4.3 for No.\ 2 also lies outside the annotation range of $[0.0,3.5]$, it is closer to an extremum, and there is much lower agreement here, suggesting that the prediction is better than that for No.\ 1.
No.\ 3 seems to be better again, as the model prediction of 0.6 is both (just) within the annotation range of $[0.5,4.5]$ and closer to the average for a similarly low-agreement instance.
Based on the standard evaluation methodology in STS research of calculating the Pearson correlation over the mean rating, however, No.\ 1 would likely be assessed as being a more accurate prediction than Nos.\ 2 or 3, based solely on how close the scalar prediction is to the annotator mean.
A more nuanced evaluation should take into consideration the relative distribution of annotator scores, and assuming a model which outputs a score distribution rather than a simple scalar, the relative fit between the two.
We return to explore this question in \Cref{sec:baselines}.

Based on these observations, we firstly study \textit{how to aggregate a collection of ratings into a representation which better reflects the ground truth}, and further go on to consider evaluation metrics which \textit{measure the fit between the distribution of annotations and score distribution of a given model}.

\subsection{Human Disagreements in Annotations}
\label{sec:humandisagreements}
\textbf{Individual Annotation Uncertainty}
Past discussions of disagreement on STS have mostly focused on uncertainty stemming from an individual annotator and the noisiness of the data collection process. 
They tend to attribute an outlier label to ``inattentive'' raters.
This has led to the design of annotation processes to control the reliability of individual ratings and achieve high inter-annotator agreement~\citep{wang2018medsts}.  
However, disagreements persist. \\
\textbf{Inherent Disagreements Among Humans}
Studies in NLI have demonstrated that disagreements among annotations are reproducible signals~\citep{pavlick-kwiatkowski-2019-inherent}.
It has also been acknowledged that disagreement is an intrinsic property of  subjective tasks~\citep{nie-etal-2020-learn, wang2022capture,plank-2022-problem}.

Despite this, most work in STS still has attributed high levels of disagreement to poor-quality data~\citep{wang2022noisy}, and has focused on reducing the uncertainty in STS modelling and providing reliable predictions~\citep{wang2022uncertaintysts}.
Little attention has been paid to analysing the inherent underlying variation in STS annotations on a continuous rating scale, or how to fit the collective human opinions to a mathematical representation.
Does a real value, Gaussian distribution, Gaussian mixture model, or a more complicated distribution  most effectively approximate the latent truth?

The shortage of individual annotator labels in STS has been a critical obstacle to in-depth analysis of disagreements among human judgements, since only the averaged similarity scores are available to the public for almost all STS datasets, apart from two small-scale biomedical benchmarks with 0.1k and 1k examples, respectively.
To this end, we first construct a large-scale STS corpus in this work with 4-19 annotators for each of almost 15k sentence pairs.
We focus on analysing disagreements among annotators 
instead of the individual uncertainty, presuming that each individual rater is attentive under a quality-controlled annotation process.

\subsection{Chinese STS Corpus}
\label{sec:chinesestscorpus}
Most progress on STS, driven by large-scale investment in datasets and advances in pre-training, has centred around English.\footnote{English STS models have achieved $r = 0.91$, while for Chinese the best results are markedly lower at $r = 0.82$ for STS-B test.} 
Efforts to build comparable datasets for other languages have largely focused on (automatically) translating existing English STS datasets~\cite{multilingualstsb, pawsx2019emnlp}. 
However, this approach may come with biases (see \secref{sec:multilingual}).
Our dataset is generated from Chinese rather than English sources, and we employ native Chinese speakers as annotators, producing the first large-scale Chinese STS dataset.\footnote{Apart from translated STS-B, there are only two Chinese corpora related to STS: BQ~\cite{chen-etal-2018-bq} and LCQMC~\cite{liu-etal-2018-lcqmc} for paraphrase detection (binary).}

%% file: section/datacollection.tex
\section{Data Collection}
\label{sec:datacollection}
We collected STS judgements from multiple annotators to estimate the distribution, for sentence pairs drawn from three multilingual sources.
\Cref{sec:datasources,sec:annotation} provide details of the collection, along with challenges in the annotation and how we ensure data quality.
All data and annotations are available at \url{https://github.com/yuxiaw/USTS}. 

\subsection{Data Sources}
\label{sec:datasources}
The first step is to gather sentence pairs.
In response to rapid rises in STS performance and insights into the shortcomings of current models and limitations of existing datasets, we create a new corpus that not only incorporates inherent human disagreements in the gold label representation, but also includes more challenging examples, on which state-of-the-art STS models tend to make wrong predictions. 
\\
\textbf{Common errors:}
our analysis over general STS-B and clinical N2C2-STS exposes three major error types.
More than half of errors lie in subsets where human agreement is low.
High uncertainty in STS labelling leads to pervasive disagreement among human judgements. 

Another is attributed to the lack of reasoning, as Nos.\,1 and 3 in \tabref{tab:examplesofvaryingdisagreements} reveal: (1) matching an abbreviation with its full name, e.g.\ \ex{Supreme Court} to \ex{SC}; and (2) building connections between descriptions that are lexically divergent but semantically related, e.g.\ \ex{carrot} and \ex{orange food}. 
The other is the failure to distinguish pairs with high lexical overlap but opposite meaning, due to word substitution or reordering.

However, these types of examples account for only a tiny proportion of existing test sets and have minimal impact on results. 
Thus, our goal is to gather more cases of high ambiguity, requiring reasoning abilities and more semantic attention in annotation.

As our data sources, we use sentences from TED talks, and sentence pairs from NLI and paraphrase corpora, as detailed below.
The combined dataset contains 14,951 pairs, which we perform basic data cleaning over to remove repeated punctuation marks (e.g.\ multiple quotation marks, dashes, or blank spaces).

\subsubsection{TED-X}
\label{sec:ted}
Compared to written texts such as essays, spoken texts are more spontaneous and typically less formal~\cite{clark2002speaking}.
Without any contextual cues such as prosody or multi-modality to help interpret utterances, readers may have trouble understanding, especially for single sentences out of context~\cite{chafe1994discourse}, resulting in high uncertainty in labelling.
We therefore choose TED speech transcriptions to gather high-ambiguity examples.

\paragraph{Selecting Single Sentences}
TED2020 contains a crawl of nearly 4000 TED and TED-X transcripts, translated into more than 100 languages. 
Sentences are aligned to create a parallel corpus~\cite{reimers-2020-multilingual-sentence-bert}.
We extracted 157,047 sentences for zh-cn with character length ranging between 20 and 100, and aligned it with the other 8 languages of en, de, es, fr, it, ja, ko, ru, and traditional zh.

\paragraph{Pairing by Retrieval}
Sentence pairs generated by random sampling are prone to be semantically distant. 
To avoid pairs with similarity scores overwhelmingly distributed in the range $[0,1]$, we use embedding-based retrieval.
For each sentence, we search for the two most similar sentences based on \textit{faiss} \cite{JDH17} using the \simcse sentence embedding of \textit{sup-simcse-bert-base-uncased}~\cite{gao-etal-2021-simcse}, obtaining 155,659 pairs after deduplication.\footnote{Note that we base this on the English versions of each sentence, due to the higher availability of pre-trained language models and sentence encoders for English.}
That is, we use (approximate) cosine similarity based on contextualised sentence embeddings instead of the surface string-based measures of previous work to sample sentence pairs.
This is expected to find pairs with a higher level of semantic overlap, rather than some minimal level of lexical match.

\begin{table*}[t!]
\centering
\resizebox{0.9\textwidth}{!}{
    \begin{tabular}{c l}
        \toprule
        Score & Description \\
        \midrule
        5 & The two sentences are completely equivalent, as they mean the same thing.\\
        4 & The two sentences are mostly equivalent, but some unimportant details differ.\\
        3 & The two sentences are roughly equivalent, but some important information differs/missing.\\
        2 & The two sentences are not equivalent, but share some details.\\
        1 & The two sentences are not equivalent, but are on the same topic.\\
        0 & The two sentences are completely dissimilar.\\
        \bottomrule
    \end{tabular}}
\caption{Similarity scores with descriptions~\cite{agirre2013sem}.}
\label{tab:annotationrule}
\end{table*}

\paragraph{Selecting Low-agreement Examples}
To select what we expect to be examples with low agreement, we leverage the observation that high-variance examples tend to be associated with low human agreement~\citep{nie-etal-2020-learn}.
That is, we keep pairs with large predictive variance, and predictions that differ greatly between two agents.

We use a \textit{bert-base-uncased}-based STS model fine-tuned on the STS-B training data  
for prediction. We obtain the mean $\mu$ and standard deviation $\sigma$ for each example from sub-networks based on \mcdropout, where $\mu$ is re-scaled to the same magnitude $[0,1]$ as the normalised $L_2$ using \simcse embedding $\mathbf{x}$, and $len_{word}(S_{en})$ is the word-level length of the English sentence.
We then select instances which satisfy the three criteria: (1) $|\frac{1}{5} \mu - (1.0 - L_2(\mathbf{x}_1, \mathbf{x}_2))| \geq 0.25$; (2) $\sigma \geq 0.16$; and (3) $len_{word}(S_{en}) \geq 12$.\footnote{We tuned these threshold values empirically, until the majority of sampled instances fell into the range $[1,3]$ --- the score interval most associated with ambiguous instances.}
This results in 9,462 sentence pairs.

\subsubsection{XNLI}
Though sentence pairs from \sickr and \unli~\citep{chen2020uncertain} are annotated with \textit{entailment} and \textit{contradiction} relations and also continuous labels, they don't specifically address semantic equivalence:
the scores in \sickr reflect semantic relatedness rather than similarity, and in \unli the annotators were asked to estimate how likely the situation described in the hypothesis sentence would be true given the premise.

We use sentence pairs from Cross-lingual NLI (\xnli: \citet{conneau2018xnli}) where there is label disagreement (which we hypothesise reflects ambiguity), noting that the dataset was annotated for textual entailment in en, and translated into 14 languages: {fr}, {es}, {de}, {el}, {bg}, {ru}, {tr}, {ar}, {vi}, {th}, {zh}, {hi}, {sw} and {ur}.
From the development (2,490) and test sets (5,010), we select examples where there is not full annotation agreement among the five annotators, resulting in 3,259 sentence pairs (1,097 dev and 2,162 test).

\subsubsection{\pawsx}
We sample 2230 sentence pairs from  \pawsx~\citep{pawsx2019emnlp} which are not paraphrases but have high lexical overlap.
Note that this is an extension of \paws~\citep{zhang-etal-2019-paws} to include six typologically-diverse languages: {fr}, {es}, {de}, {zh}, {ja} and {ko}.

\subsection{Annotation}
\label{sec:annotation}
We employ four professional human annotators (all Chinese native speakers) to assign labels to the 14,951 Chinese sentence pairs in the first round, and an additional 15 annotators to provide additional annotations for 6,051 examples of low human agreement (as detailed below).

\paragraph{Annotation Guideline} 
\tabref{tab:annotationrule} shows the 6-point ordinal similarity scale we use, plus definitions.

\paragraph{Quality Control}
It is difficult to ensure that any divergences in annotations are more likely due to task subjectivity or language ambiguity than inattentiveness.
We attempt to achieve this by not using crowdsourced workers, but instead training up in-house professional annotators with expert-level knowledge in Linguistics, and significant experience in data labelling.
They were first required to study the annotation guidelines and exemplars, and then asked to annotate up to 15 instances of high-agreement pre-selected from the STS-B training set.
For each example, the annotation is regarded to be correct when the difference between the assigned and gold-standard label is $<$0.5.
Failing this, the annotator is provided with the correct label and asked to annotated another instance.

This procedure was iterated for three rounds to familiarize the annotators with the task.
On completion of the training, we only retain annotators who achieve a cumulative accuracy of $\ge$75\%.

\subsection{Analysis of First-round Annotations}
\label{sec:firstroundanalysis}
\paragraph{Dataset breakdown} \tabref{tab:corpus} shows the breakdown of instances across the three component sets, as well as the combined \usts dataset.
In terms of average length (\textit{zh} character level), \xnli is the shortest on average (esp.\ for S2, the hypothesis), followed by \tedx and \pawsx. 

\paragraph{Inter-annotator agreement}
The average Pearson ($r$) and Spearman ($\rho$) correlation between the six pairings of annotators, and standard deviation ($\sigma$) among the four annotators, are $r=0.74$, $\rho=0.68$, $\sigma=0.47$.
These numbers reflect the fact that there is high disagreement for a substantial number of instances in \usts, in line with the sampling criteria used to construct the dataset. 
As such, aggregating ratings by \textit{averaging} is not able to capture the true nature of much of the data.
Two questions naturally arise: 
(1) at what level of variance does averaging noticeably bias the gold label? and (2) how should annotations be aggregated to fit the latent truth most closely?

\begin{table}[t!]
\centering
\resizebox{\columnwidth}{!}{
    \begin{tabular}{l cccc}
        \toprule
        Source & \tedx & \xnli & \pawsx & \usts \\
        \midrule
        \textbf{Amount} &&&& \\
        raw              & 9462 & 3259 & 2230 & 14951 \\
        $\sigma>0.5$     & 3458 & 1597 & \z996  & 6051 \\
        ratio            & 36.5\% & 49.0\%& 44.7\% & 40.5\% \\ 
        \midrule
        \textbf{Length} &&&& \\
        S1   & 39.0 & 34.0 & 43.5 & 38.6 \\
        S2   & 39.2 & 16.9 & 43.3 & 34.9 \\
        pair & 39.1 & 25.4 & 43.4 & 36.8 \\
        \midrule
        \textbf{Raters}& & & & \\
        $r$      & 0.48 & 0.61 & 0.49 & 0.74 \\
        $\rho$   & 0.50 & 0.58 & 0.41 & 0.68 \\
        $\sigma$ & 0.44 & 0.52 & 0.49 & 0.47 \\
        \midrule
        \textbf{STSb-zh} & & & &  \\
        $r$      & 0.41 & 0.48 & 0.32 & 0.70 \\
        $\rho$   & 0.43 & 0.50 & 0.18 & 0.63 \\
        $\sigma$ & 0.21 & 0.22 & 0.19 & 0.21 \\
        \bottomrule
    \end{tabular} }
  \caption{Details of the \usts dataset. ``$r$'' = Pearson's correlation; ``$\rho$'' = Spearman's rank correlation; and ``$\sigma$'' = standard deviation}
\label{tab:corpus}
\end{table}

\begin{figure}[t!]
	\centering
	\includegraphics[scale=0.5]{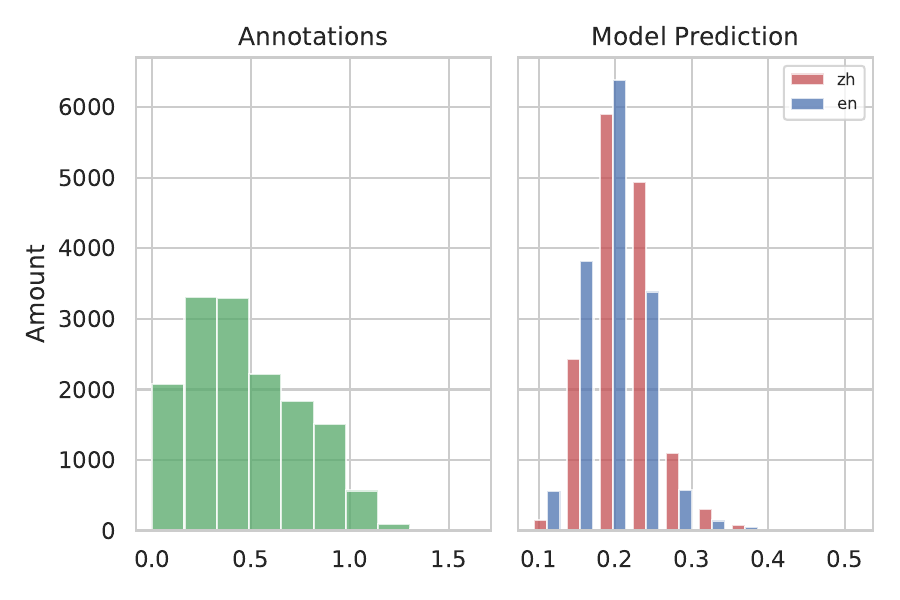}
	\caption{Standard deviation distribution of the four first-stage annotators (left) and model predictions (right).}
	\label{fig:stddistribution}
\end{figure}

\paragraph{High vs.\ Low agreement}
\figref{fig:stddistribution} shows the first-round variance distribution, wherein $\sigma$ ranges from 0.0 to 1.5, with 8,900 pairs being less than 0.5.
It indicates that on $\sim$60\% examples, the assessments of four annotators fluctuate around the average score in a smaller range (0.0--0.5 on average), while the judgements of the remaining 6,051 pairs are spread out over a wider range (0.5--1.5).

We sample 100 examples and find that, when $\sigma \leq$ 0.5, generally more than 10 out of 15 annotators highly agree with each other.
This basically satisfies the assumption that makes \textit{averaging} less biased: \textit{individual ratings do not vary significantly}~\citep{lee2005empirical}.
While less than half annotators reach consensus when $\sigma>$ 0.7, and less than 5 when $\sigma \geq$ 1.0 (referring back to our earlier examples in \tabref{tab:examplesofvaryingdisagreements}).
Thus, we heuristically regard $\sigma$=0.5 as a tipping point for distinguishing examples of low ($\sigma>0.5$) and high agreement ($\sigma \leq0.5$).

Accordingly, we split the data into two subsets, reflecting the different levels of disagreement: cases where $\sigma\le0.5$ are \textit{uncontroversial} (\textbf{\ustsu}); and cases where $\sigma>0.5$  are \textit{contentious} (\textbf{\ustsc}).

\paragraph{Does the model agree with the annotators?}
We take \textit{bert-base-chinese} and fine-tune it on the Chinese STS-B training data\footnote{Chinese STS-B has 5,231, 1,458 and 1,361 examples for training, validation and test, respectively; see \url{https://github.com/pluto-junzeng/CNSD}} with a learning rate of 2e-5 for 3 epochs, obtaining $r{=}0.82$/$\rho{=}0.82$ on the validation set, and $r{=}0.80$/$\rho{=}0.79$ on the test set; we refer to this model as ``STSb-zh''.
We compute $r$ and $\rho$ between the model prediction and each of the four annotations, and present the average results in \tabref{tab:corpus}.

Both $r$ and $\rho$ across \tedx, \xnli, and \pawsx are below 0.5, with \pawsx being particularly bad with half of the pairs being predicted to be in the range $[4,5]$.
Predictions of \usts are primarily concentrated in the range $[1,3]$, when majority annotations are in the range $[0,2]$

This suggests it is non-trivial for current models to perform well without training on \usts, and models tend to over-assign high scores (\figref{fig:stddistribution}: predictive $\sigma$ is $<0.3$ vs.\ annotator $\hat{\sigma}=0.47$).
However, it also leads us to consider whether the distribution estimated based on the four annotators is adequate to generate a gold standard. 
To this end, we investigate the question \textit{How does the collective distribution vary when increasing the number of annotators, on cases of uncontroversial \ustsu and contentious \ustsc?}

\subsection{Collective Distribution Analysis}
\label{sec:secondcasestudy}
We measure the distributional variation through 
(1) fluctuation of $\mu$ and $\sigma$; and
(2) distributional divergence between first-round and second-round annotators.

\textbf{Study design:} we sample 100 instances from \ustsu and 100 from \ustsc, with a ratio of 4:3:3 from \tedx, \xnli, and \pawsx, resp. 
We then had another 15 qualified Chinese native annotators score the 200 Chinese sentence pairs.

Formally, the annotation matrix $A^{N \times M}$ represents a data set with $N$ examples annotated by $M$ annotators.
In our setting, $N{=}100$ and $M{=}19$ for both \ustsu and \ustsc.
We capture the variation of $\mu$ and $\sigma$ over 100 examples by averaging $\boldsymbol{\mu}${=}mean(A[:,:i], axis{=}1) and $\boldsymbol{\sigma}${=}std(A[:,:i], axis{=}1), where $i$ ranges from 4 to 19, incorporating the new ratings incrementally.

The collective distribution for the first-round annotation A[:,:4] is denoted as $\mathbf{p}{=}\mathcal{N}(\boldsymbol{\mu_1}, \boldsymbol{\sigma_1})$, and $\mathbf{q}{=}\mathcal{N}(\boldsymbol{\mu_2}, \boldsymbol{\sigma_2})$ for A[:,4:4+j] as we add new annotators.
We observe the KL-Divergence$(p\|q)$ as we increase $j$.

\textbf{Hypothesis:}
We hypothesise that the distribution will remain stable regardless of the number of annotators on the uncontroversial \ustsu, but change substantially on the contentious \ustsc. 

\begin{figure}[t!]
	\centering
	\includegraphics[scale=0.5]{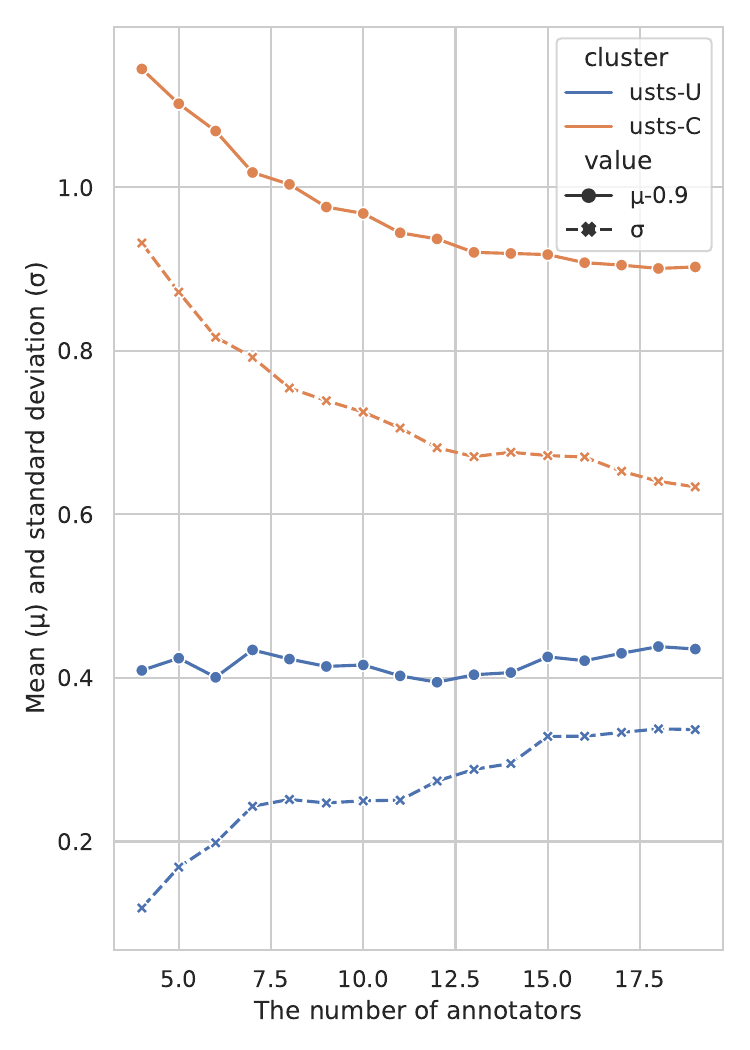}
	\caption{Average $\mu$ and $\sigma$ over 100 examples of \ustsu and \ustsc as we incorporate new annotators.}
	\label{fig:mustdline}
\end{figure}

\textbf{Results:} To plot the value of $\mu$ and $\sigma$ in the same figure, we re-scale $\mu$ by subtracting $0.9$ in \figref{fig:mustdline}. 
We find that with an increased number of annotators, $\mu$ of \ustsu remains stable with minor perturbations, while $\mu$ of \ustsc declines and steadily flattens out.

On \ustsu, $\sigma$ ascends slowly and converges to 0.3. This matches our expectation that increasing annotators will result in more variance. Yet it still varies in the range $[0.1,0.3]$ due to the high certainty of the uncontroversial examples. 
In contrast, $\sigma$ of \ustsc stays consistently high, indicating that there are still strong disagreements even with more annotators, because of the inherent ambiguity of contentious cases.
It fluctuates in a larger range of $[0.6,1.0]$, with a steeper drop. 
That is, combining more ratings results in large variations in $\mu$ and $\sigma$ for \ustsc, but less 
for \ustsu.

Therefore, the distribution obtained from four annotators is adequate for uncontroversial examples, but insufficient for \ustsc: more annotators are needed to gain a representative distribution.

\paragraph{How many annotators should be employed?}
In \figref{fig:mustdline}, $\mu$ and $\sigma$ of \ustsc vary substantially before $M{=}15$, then stabilise.
The trend of KL-Divergence in \tabref{tab:casestudyKL} demonstrates the same phenomenon: KL declines as the number of annotators increases, with a relatively small and stable divergence when $j>10$.
Combining these two, we employ 15 extra annotators to score the 6,051 cases for \ustsc in the second-round annotation.

\begin{table}[t]
\centering
\resizebox{\columnwidth}{!}{
    \begin{tabular}{l c c c c c c}
        \toprule
        $j$      & 4 & 6 & 8 & 10 & 14 & 15 \\
        \midrule
        \ustsu & \z4.26  & 2.58 & 0.98 & 1.03 & 0.91 & 0.93 \\
        \ustsc & 12.83 & 5.08 & 5.45 & 3.51 & 2.99 & 2.82 \\
        \bottomrule
    \end{tabular} }
\caption{KL-Divergence between the first-round distribution and the second-round, for increasing $j$.}
\label{tab:casestudyKL}
\end{table}

\begin{table}[t]
\centering
\resizebox{\columnwidth}{!}{
    \begin{tabular}{l c c c c c c}
        \toprule
       & \# & Annotators & $\mu$ & $\sigma$ & $r$ & $\rho$ \\
        \midrule
        STS-B  & 8,085 & 5 & 0.0--5.0 & -- & -- & -- \\
        \midrule
        \ustsu & 8,900 & 4  & 0.0--5.0 & 0.27 & 0.91 & 0.73 \\
        \ustsc & 6,051 & 19 & 0.2--4.4 & 0.56 & 0.72 & 0.63 \\
        \bottomrule
    \end{tabular} }
\caption{Statistical breakdown of STS-B (zh) and \ustsu/\ustsc; $\mu${=} the range of integrated score.}
\label{tab:ustsavsustst}
\end{table}

\paragraph{First-round vs.\ second-round:} 
\begin{figure}[t!]
	\centering
	\includegraphics[scale=0.35]{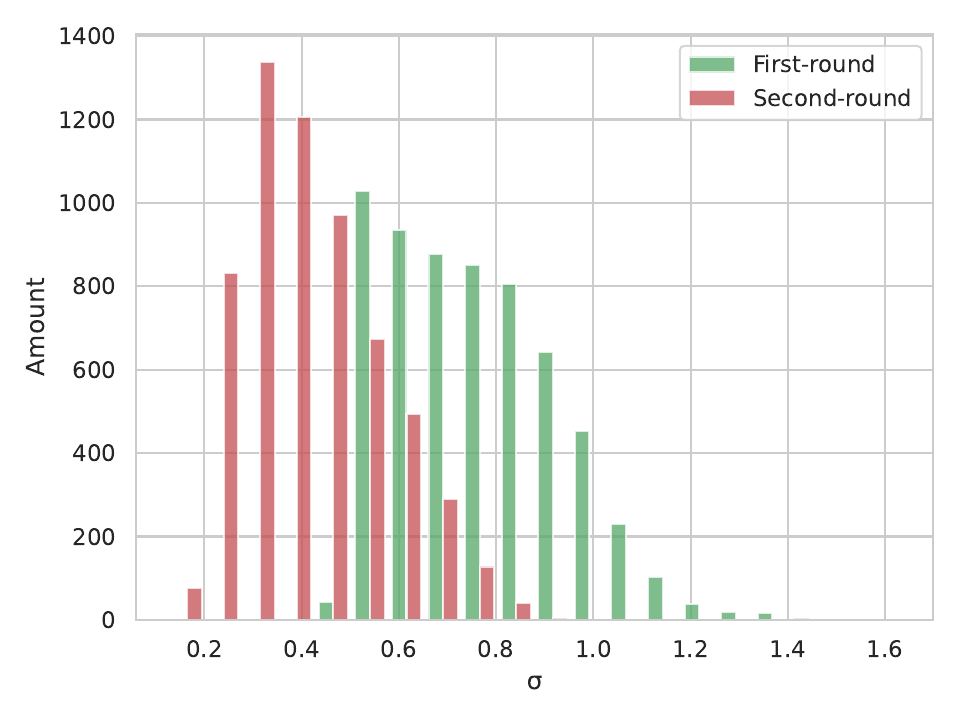}
	\includegraphics[scale=0.35]{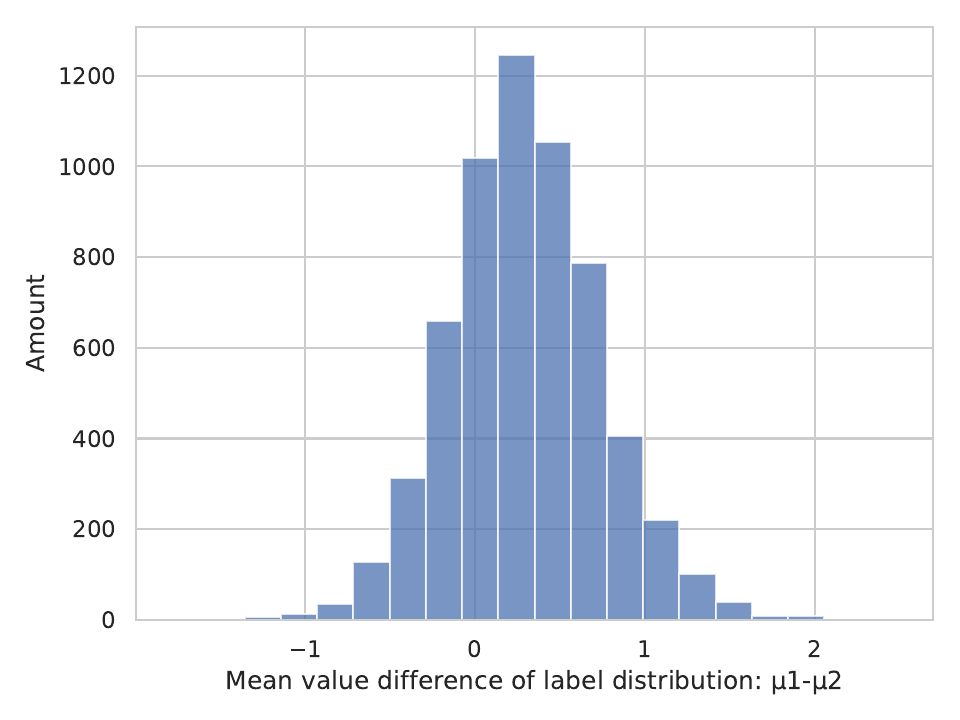}
	\caption{Distribution of $\sigma$ (top) and $\mu_1 - \mu_2$ (bottom) of the first- and second-round annotation distribution.}
	\label{fig:secondround}
\end{figure}
We compare $\sigma$ and $\mu$ between the first-round (in green) and second-round (in red) annotations in \figref{fig:secondround} (top).
The shape of the $\sigma$ distributions is very similar, but the green bars ($\sigma_1$) move towards the right by 0.3 or so, with respect to the red bars ($\sigma_2$), leading to the average $\hat{\sigma_2}${=}0.42 $\ll$ $\hat{\sigma_1}${=}0.76.
This indicates that the second-round distribution is more stable, with less overall variance.
Nonetheless, 87\% of pairs exceed the average deviation of 0.27 for \ustsu, reflecting the higher number of disagreements.
Additionally, the distribution of $\mu_1-\mu_2$ in \figref{fig:secondround} (bottom) is close to a normal distribution, within the range of $[-1,2]$. The majority are to the right of zero,  indicating that annotators in the first round tend to assign higher scores than in the second, resulting in a larger $\mu$.

\subsection{The Resulting Corpus}
\textbf{\ustsu vs.\ \ustsc} The number of examples in \ustsu and \ustsc is 8,900 and 6,051, respectively, with largely comparable $\mu$ range of $[0,5]$ and $[0.2,4.4]$ (see \tabref{tab:ustsavsustst}).
\ustsu has a much smaller $\hat{\sigma}$ of 0.27 than \ustsc ($\hat{\sigma} {=} 0.56$), consistent with their inherent uncertainty level.
Analogously, \ustsu has a higher correlation of $r{=}0.91$ among annotators, compared to $r{=}0.72$ for \ustsc.

%% file: section/humanjudgementanalysis.tex
\section{Aggregation of Human Judgements}
\label{sec:humanjudgementsaggregation}
For the high-agreement cases of \ustsu, gold labels can be approximated by aggregating multiple annotations into either a scalar or a single Gaussian distribution.
However, for low-agreement examples, how to aggregate the human ratings remains an open question. 

\begin{figure}[t!]
	\centering
	\includegraphics[scale=0.37]{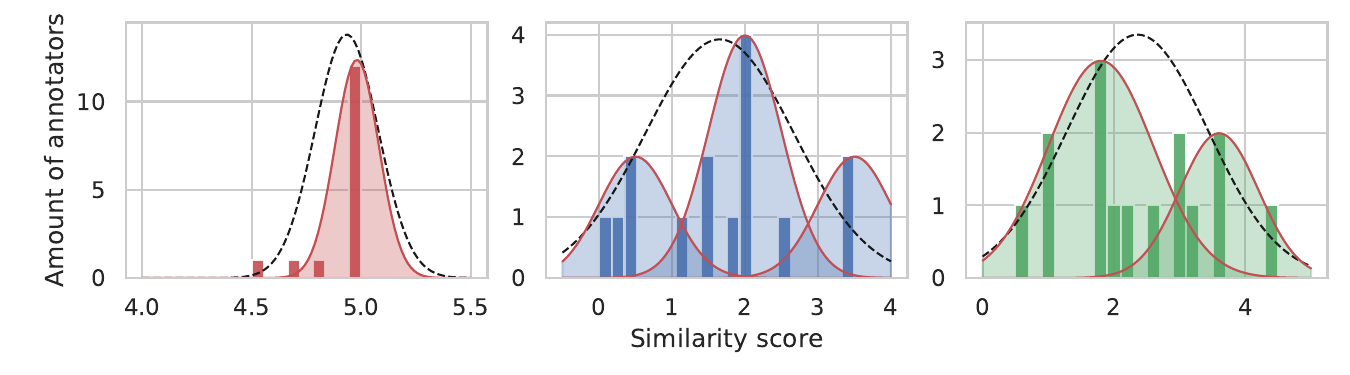}
	\caption{Human judgement distributions of examples in \tabref{tab:examplesofvaryingdisagreements}, with uni-, tri- and bi-modal Gaussian resp. The dotted black line shows the model fit when using a single Gaussian; the shaded curve shows the model learned when allowed to fit $k$ components of a GMM.}
	\label{fig:multimodaldistributions}
\end{figure}

\textbf{Are all distributions unimodal Gaussian?}
Though most distributions of human assessments can be assumed to be sampled from an underlying (generative) distribution defined by a single Gaussian, we observed judgements that a unimodal Gaussian struggles to fit.
The annotations of examples Nos.\,2 and 3  in \figref{fig:multimodaldistributions} exhibit clear bi- or tri-modal distributions. 
How often, then, and to what extent do multimodal distributions fit better?

We answer this question by fitting human judgements using a Gaussian Mixture Model (GMM), where the number of components is selected during training.
This means the model can still choose to fit the distribution with only one Gaussian component where appropriate.
If additional components yield a better fit to the judgements, i.e.\ larger log likelihood is observed than using a unimodal distribution, we consider the human judgements to exhibit a multimodal distribution.

\begin{figure}[t!]
    \centering
    \includegraphics[scale=0.37]{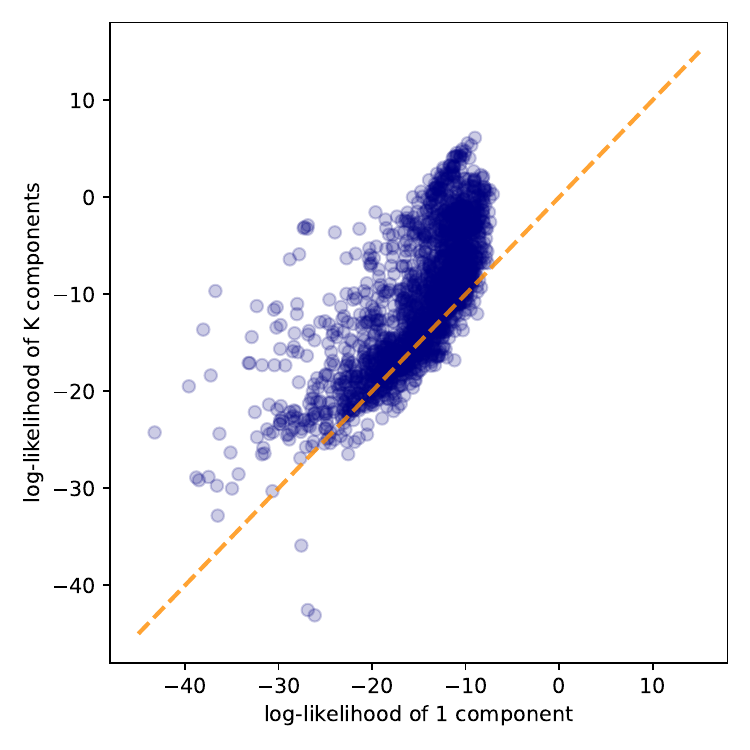}
    \includegraphics[scale=0.37]{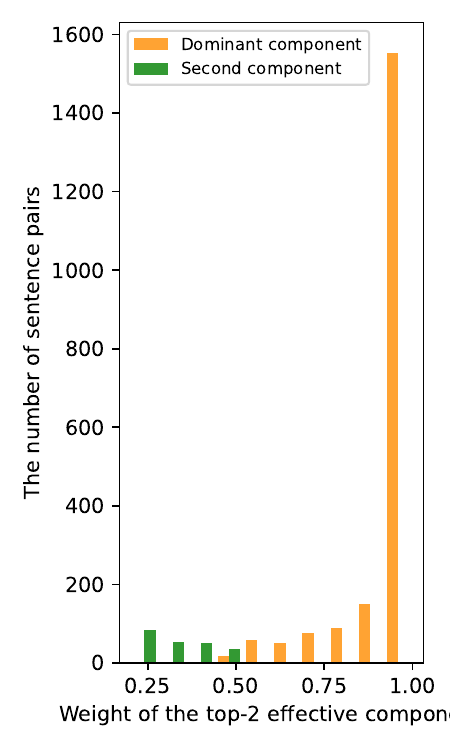}
    \caption{\textit{Left:} Log likelihood of test data under the single-component Gaussian ($x$-axis) vs.\ the $k$-component GMM ($y$-axis). The darker the area, the more the examples concentrate. \textit{Right:} Weights of top-2 effective component distribution.}
    \label{fig:testloglikelihood1vsk}
\end{figure}

\textbf{Experiments and Results}
We randomly split \ustsc into a training (4,051) and test set (2,000), and use the training data to fit a GMM with: (1) one component; or (2) the optimal number of components $k$.
We compute the log likelihood assigned to each example in the test set in \figref{fig:testloglikelihood1vsk}(left), with the unimodal results as the $x$-axis and multimodal Gaussian as the $y$-axis.
The majority of points fall on or above the diagonal line ($y=x$), with a multimodal distribution outperforming a unimodal Gaussian distribution for 83\% of instances.
However, does this suggest that most examples exhibit multiple peaks?

\textbf{Effective components:} We count the effective components for each sentence pair based on the weight assigned by the GMM in form of a probability for each component.
We see that, for 11.3\% of pairs, there is a nontrivial second component (weight$\ge0.2$), and a third component on 3 pairs. Rarely are there more than three components with significant weights (see \tabref{tab:effectivecomponents}).
Moreover, we find that the weight of the dominant component mostly (87\%) distributes over 0.8, and that the weight of the second effective component scatters across the range 0.25--0.5 (the right of \figref{fig:testloglikelihood1vsk}).
This reveals that the GMM does not frequently use more than one effective component, with much lower weights on the second or third components.
The majority of held-out human judgements fit a unimodal distribution well.
\begin{table}[!t]
\centering
\resizebox{0.9\columnwidth}{!}{
    \begin{tabular}{c r@{\,\,\,}r@{\,\,\,}r@{\,\,\,}r r r@{\,\,\,}r@{\,\,\,}r@{\,\,\,}r}
      \toprule
       & \multicolumn{3}{c }{\textbf{Testing}} & & \multicolumn{3}{c }{\textbf{Train}}\\
       K & amount & prop(\%) & $\hat{\sigma}$ & & amount & prop & $\hat{\sigma}$ \\ 
      \cmidrule{2-4}
      \cmidrule{6-8}
      1 & 1772 & 88.6 & 0.55 & & 3755 & 92.7 & 0.48 \\
      2 & 225  & 11.3 & 0.63 & & 294  & 7.3  & 0.50 \\
      3 & 3    & 0.0  & 0.39 & & 2    & 0.0  & 0.66 \\
      \bottomrule
    \end{tabular}}
    \caption{The amount and averaged standard deviation $\hat{\sigma}$ of examples with $k=\{1,2,3\}$ effective components of human judgements distributions in the training and test splits.}
    \label{tab:effectivecomponents}
\end{table}

\textbf{Gold Labels:} Given that a minority of instances in \ustsc are bimodally distributed, and that even for these instances, the weight on the second components is low, we conservatively use a single Gaussian to aggregate human judgements for all cases in this work.

%% file: section/prediction.tex
\section{Analysis of Model Predictions}
\label{sec:baselines}
Most STS models predict a pointwise similarity score rather than of a distribution over values. 
\citet{wang2022uncertaintysts} estimated the uncertainty for continuous labels by \mcdropout and Gaussian process regression (GPR).
However, due to the lack of gold distributions, they only evaluate outputs using expected calibration error (\ece) and negative log-probability density (\nlpd), assessing the predictive reliability.
It's unknown whether these uncertainty-aware models mimic human disagreements, i.e.\ the predicted deviation reflects the variance of human judgements.

To explore this, we experiment over \usts and incorporate distributional divergence (i.e.\ Kullback-Leibler Divergence; ``\kl'') into the evaluation, to observe the fit between the distribution of collective human judgements and the model predictive probability. 
We also examine the ability of different models to capture the averaged score for low-agreement cases, and whether a well-calibrated model fits the distribution of annotations better.

\paragraph{Evaluation Metrics:} For singular values, STS accuracy is generally evaluated with Pearson correlation ($r$) and Spearman rank correlation ($\rho$), measuring the linear correlation between model outputs and the average annotation, the degree of monotonicity under ranking, respectively.

For uncertainty-aware outputs, \ece and \nlpd can be used to assess model reliability in the absence of gold distributions.
\ece measures whether the estimated predictive confidence is aligned with the empirical correctness likelihoods.
A well-calibrated model should be less confident on erroneous predictions and more confident on correct ones.
\nlpd penalises over-confidence more strongly through logarithmic scaling, favouring under-confident models.

\input{section/tableofprediction.tex}
\subsection{Models and Setup}
\paragraph{BERT with Two-layer MLP:} 
The hidden state $\mathbf{h}$ from the last-layer hidden state of BERT \textit{CLS} token~\cite{devlin2018bert} is passed
through a two-layer MLP with $\tanh$ activation function.
We refer to this model as \textit{BERT-lr} when making deterministic predictions, and \textit{BERT-lr-MC} when using \mcdropout~\cite{gal2016dropout} for uncertainty estimation.

\paragraph{SBERT with GPR:} 
In contrast with end-to-end training, sparse GPR is applied to estimate distributions, taking encoded sentences from Sentence-BERT (SBERT: \citet{reimers-gurevych-2019-sentence}) as input.
We also calculate the cosine similarity between S1 and S2 using SBERT, as a non-Bayesian counterpart.

\paragraph{Setup:}
\texttt{\small{bert-base-chinese}} is used with input format \texttt{\small[CLS] S1 [SEP] S2 [SEP]} for text pair $(S1, S2)$, implemented based on the \textit{hugging-face Transformer} framework.
We fine-tune SBERT separately over each STS corpus based on \texttt{\small{bert-base-chinese-nli}}, using the same configuration as the original paper.
We use the concatenation of the embeddings $u \oplus v$, along with their absolute difference $|u-v|$ and element-wise multiplication $v \times t$ to represent a sentence pair, implemented in pyro.\footnote{\url{https://pyro.ai/}}

We evaluate \stsb, \ustsu, and \ustsc under five training settings, as presented in \tabref{tab:baseline}:
\begin{compactenum}
\item Zero-shot: SBERT with no tuning;
\item GPR trained on \texttt{\small{sbert-nli}};
\item Domain-specific: fine-tuned on each dataset separately;
\item Domain-generalised: fine-tuned using the three datasets combined;
\item Cross-domain: train with \stsb training data for \ustsu and \ustsc, and with \usts for \stsb.
\end{compactenum}

\subsection{Results and Analysis}
\textbf{\usts is challenging.}
In setting (1) of \tabref{tab:baseline}, purely depending on pre-trained semantic representation and cosine similarity, correlations over \ustsu and \ustsc are much lower than \stsb.
This suggests that \usts is a challenging dataset, but can be learned.
\ustsu in particular achieves large improvements in performance after domain-specific training in experiments (3)--(4).

\begin{figure}[t!]
    \centering
    \includegraphics[scale=0.55]{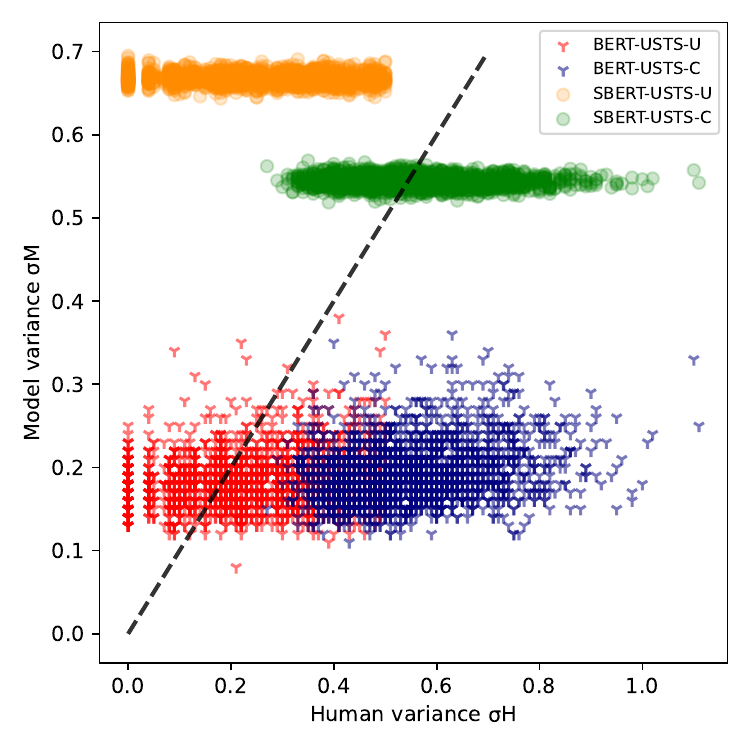} 
    \caption{Predicted variance $\sigma_M$ ($y$-axis) with increasing human disagreement ($x$-axis). The Red and blue triangles = \ustsu and \ustsc from experiment setting (4) in \tabref{tab:baseline}, orange and green circles = \ustsu and \ustsc from experiment setting (5), and black line is $y=x$. \ustsu disperses at the left of the $x$-axis and low-agreement \ustsc scatters to the right.}\label{fig:testscattersigmaovertwomodels}
\end{figure}

\textbf{Critical differences exist between model outputs and human annotations.}
The models can capture average opinion, resulting in reasonable $r$/$\rho$ between the predicted target value and the averaged annotations. However, they cannot capture the variance of human opinions.
To quantify how well the predicted variance $\sigma_M$ captures the variance $\sigma_H$ of human judgements, we analyse the outputs of the top-2 settings: BERT-lr-MC from setting (4) and SBERT-GPR from setting (5), for \ustsu and \ustsc.
We compute the correlation $r$ and $\rho$ between $\boldsymbol{\sigma}_M$ and $\boldsymbol{\sigma}_H$ in \tabref{tab:sigmacorrelation}, and visualise the $\sigma_M$ with increasing human disagreement in \figref{fig:testscattersigmaovertwomodels}.

\begin{table}[!t]
\centering
\resizebox{0.9\columnwidth}{!}{
    \begin{tabular}{l r@{\,\,\,}r@{\,\,\,}r@{\,\,\,}r r r@{\,\,\,}r@{\,\,\,}r@{\,\,\,}r}
      \toprule
       & \multicolumn{3}{c }{\textbf{\ustsu($\hat{\sigma_H}$=0.26)}} & & \multicolumn{3}{c }{\textbf{\ustsc($\hat{\sigma_H}$=0.56)}}\\
       Model & $r$ & $\rho$ & $\hat{\sigma_M}$ & & $r$ & $\rho$ & $\hat{\sigma_M}$ \\ 
      \cmidrule{2-4}
      \cmidrule{6-8}
      (4) BERT-lr-MC & 0.13 & 0.12 & 0.19 & & 0.24 & 0.23 & 0.20 \\
      (5) SBERT-GPR & $-$0.07  & $-$0.06 & 0.67 & & $-$0.05  & $-$0.06  & 0.54 \\
      \bottomrule
    \end{tabular}}
    \caption{Test set correlation between the predicted variance and collective human variance.}
    \label{tab:sigmacorrelation}
\end{table}

There is no apparent correlation between $\sigma_M$ and $\sigma_H$. 
A given model displays similar deviation $\sigma_M$ regardless of the relative amount of human disagreement. Different models concentrate on different parts of the spectrum, e.g.\ BERT-lr-MC is distributed in the range $[0.1,0.2]$ while SBERT-GPR is distributed in the range $[0.5,0.7]$, and neither follows the line of $y=x$. This suggests that \textbf{the uncertainty captured by current models is not the uncertainty underlying human disagreements}. Rather it may \textbf{reflect the model’s predictive confidence} on the data set as a whole. 
This finding is not surprising since none of the models are optimised to capture collective human opinions, but suggests an important direction for future improvement.

\textbf{Being trustworthy is orthogonal to being accurate.}
We see that \ece and \nlpd do not mirror the results for $r$/$\rho$ and distributional divergence \kl.
This implies the ability required to improve model reliability differs from that required to perform accurately, regardless of whether a target value or a target distribution is predicted.

\textbf{Low human-agreement \usts is detrimental to training sentence embeddings.}
Comparing the performance of experiment settings (2) and (5) in \tabref{tab:baseline}, tuning SBERT on \usts hurts results over \stsb across the board, while training on \stsb benefits both \ustsu and \ustsc.
We speculate that the examples in \usts with larger annotator variance are more ambiguous than \stsb.
Forcing networks to learn from high-ambiguity signals may inhibit generalisation, resulting in worse representations.

\textbf{Discussion} For instances of high disagreement, neither a scalar nor a single Gaussian fits a set of observed judgements adequately. As a direction for future work, we suggest exploring the direct estimation of individual ratings (e.g.\ by few-shot prompt-based prediction) and evaluating against the raw collective opinions.
This could circumvent the ineffective training and evaluation caused by aggregation.

%% file: section/tableofprediction.tex
\begin{table*}[!t]
\centering
\resizebox{\textwidth}{!}{
    \begin{tabular}{l l r@{\,\,\,}r@{\,\,\,}r@{\,\,\,}r r r@{\,\,\,}r@{\,\,\,}r@{\,\,\,}r@{\,\,\,}r r r@{\,\,\,}r@{\,\,\,}r@{\,\,\,}r@{\,\,\,}r}
        \toprule
        & & \multicolumn{4}{c }{\textbf{\stsb}} & & \multicolumn{5}{c }{\textbf{\ustsu}}
        & & \multicolumn{5}{c }{\textbf{\ustsc}}\\
      \cmidrule{3-6}
      \cmidrule{8-12}
      \cmidrule{14-18}
      \textbf{} & \textbf{Model} &
      $r$ $\uparrow$ & $\rho$ $\uparrow$ & \ece $\downarrow$ & \nlpd $\downarrow$ & &
      $r$ $\uparrow$ & $\rho$ $\uparrow$ & \ece $\downarrow$ & \nlpd $\downarrow$ & \kl $\downarrow$ & & 
      $r$ $\uparrow$ & $\rho$ $\uparrow$ & \ece $\downarrow$ & \nlpd $\downarrow$ & \kl $\downarrow$ \\
      \midrule
      \multicolumn{6}{l}{\textbf{\textit{SBERT-NLI}}} \\
      \textbf{(1)} & SBERT-cosine & 0.714 & 0.718 & \NA    & \NA    & & 0.597 & 0.383 & \NA & \NA & \NA & & 0.572 & 0.442 & \NA & \NA & \NA \\
      \textbf{(2)} & SBERT-GPR    & 0.741 & 0.743 & \underline{0.001} & \underline{0.532} & & 0.709 & 0.433 & \underline{0.020} & 0.033 & 2.233 & & 0.656 & 0.455 & \underline{0.139} & $-$0.09 & 0.576 \\
      \midrule
      \multicolumn{6}{l}{\textbf{\textit{Domain-specific}}} \\
      \multirow{4}{*}{\textbf{(3)}} 
      & BERT-lr &    0.808 & 0.804 & \NA & \NA & & 0.855 & 0.700 & \NA & \NA & \NA & & 0.806 & 0.707 & \NA & \NA & \NA \\
      & BERT-lr-MC & 0.811 & 0.805 & \textbf{0.167} & \textbf{4.709} & & 0.856 & \textbf{0.703} & \textbf{0.054} & 1.079 & 4.587 & & 0.809 & 0.708 & \textbf{0.046} & \textbf{0.442} & 6.073\\
      & SBERT-cosine & 0.779 & 0.781 & \NA & \NA & & 0.661 & 0.387 & \NA & \NA & \NA & & 0.596 & 0.460 & \NA & \NA & \NA \\
      & SBERT-GPR & \underline{0.780} & \underline{0.782} & 0.053 & 0.917 & & 
      0.683 & 0.388 & 0.137 & 0.651 & 3.050  & & 0.606 & 0.444 & 0.415 & 0.717 & 0.950 \\
      \midrule
      \multicolumn{6}{l}{\textbf{\textit{Domain-generalised}}} \\
      \multirow{4}{*}{\textbf{(4)}}
      & BERT-lr & \textbf{0.815} & \textbf{0.813} & \NA & \NA & & 0.860 & 0.692 & \NA & \NA & \NA & & 0.835 & 0.768 & \NA & \NA & \NA \\
      & BERT-lr-MC & 0.814 & 0.811 & 0.179 & 5.865 & & \textbf{0.861} & 0.697 & 0.060 & \textbf{0.898} & \textbf{4.434} & & \textbf{0.838} & \textbf{0.774} & 0.278 & 0.702 & \textbf{5.401} \\
      & SBERT-cosine & 0.772 & 0.772 & \NA & \NA & &  0.686 & 0.435 & \NA & \NA & \NA & & 0.670 & \underline{0.523} & \NA & \NA & \NA \\
      & SBERT-GPR & 0.772 & 0.775 & 0.017 & 0.645 & & 0.707 & 0.433 & 0.098 & 0.268 & 2.578 & & 0.674 & 0.497 & 0.157 & \underline{$-$0.04} & 0.955 \\
      \midrule
      \midrule
     \multicolumn{6}{l}{\textbf{\textit{Cross-domain}}} \\
     \multirow{4}{*}{\textbf{(5)}} 
     & BERT-lr & 0.675 & 0.667 & \NA & \NA & & 0.754 & 0.650 & \NA & \NA & \NA & & 0.725 & 0.676 & \NA & \NA & \NA  \\
      & BERT-lr-MC & 0.678 & 0.671 & 0.348 & 12.90 & & 0.755 & 0.695 & 1.296 & 10.55 & 13.95 & & 0.729 & 0.687 & 1.298 & 8.956 & 12.62 \\
      & SBERT-cosine & 0.695 & 0.692 & \NA & \NA & & 0.647 & 0.449 & \NA & \NA & \NA & & 0.606 & 0.481 & \NA & \NA & \NA \\
      & SBERT-GPR & 0.726 & 0.726 & 0.001 & 0.555 & &  
      \underline{0.723} & \underline{0.481} & \underline{0.020} & \underline{0.012} & \underline{2.215} & & \underline{0.675} & 0.494 & 0.148 & $-$0.11 & \underline{0.555} \\
        \bottomrule
    \end{tabular}}
    \caption{Test set correlation ($r$/$\rho$), \ece, \nlpd and \kl
      using end-to-end (BERT) and pipeline (SBERT), over \stsb, \ustsu
      and \ustsc, under five settings. The bold number is the best
      result for BERT, and the underlined number is that for SBERT.}
    \label{tab:baseline}
  \end{table*}

%% file: section/multilingual.tex
\section{Multilingual USTS}
\label{sec:multilingual}
\begin{figure*}[t!]
	\centering
	\includegraphics[scale=0.5]{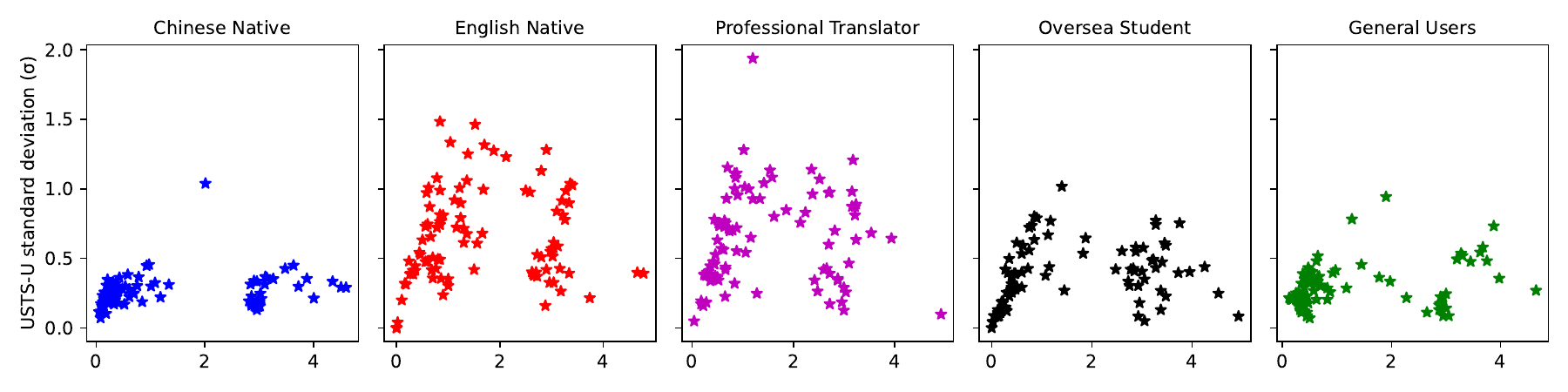} \\
	\includegraphics[scale=0.5]{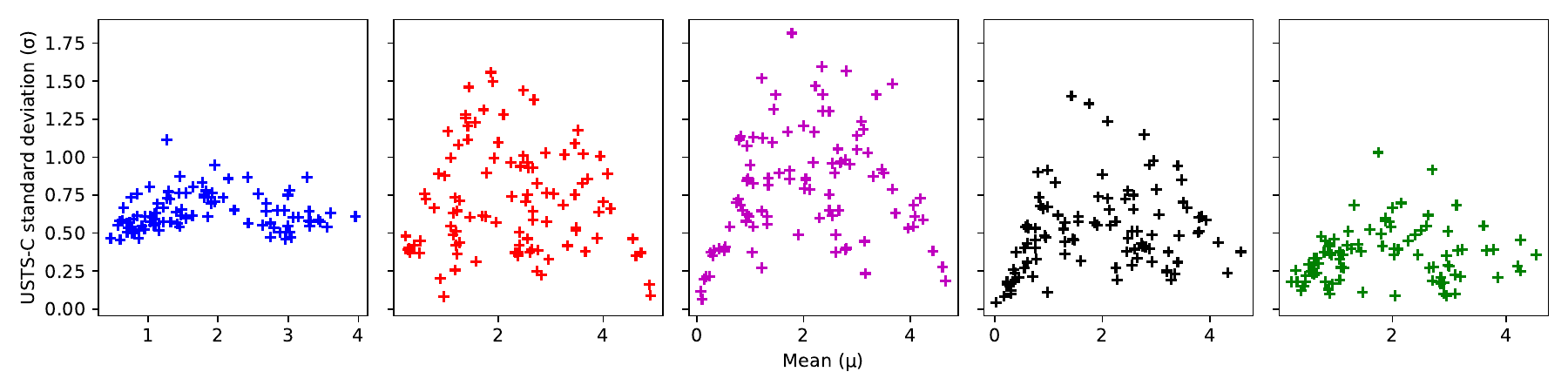}
	\caption{Scatter plot of 100 examples sampled from \ustsu (top) and \ustsc (bottom) annotated by Chinese native, English native, professional translator (PT), overseas students (OS), and general users (GU). We plot ($\mu$,$\sigma$) as coordinate points.}
	\label{fig:multilingualroles}
\end{figure*}

Before extending \usts into a multilingual benchmark, we question the validity of previous approaches involving direct transfer of annotations collected for one language to other languages~\cite{liu-etal-2021-learning-domain, pawsx2019emnlp}.
This strategy assumes that the nuanced semantics of the component sentences is not changed under translation, and hence the label will be identical.
To test whether this assumption is reasonable, we analyse the impact of language on the annotations, and 
discuss whether such ratings are transferable  across languages.

Specifically, we establish whether the label distribution varies based on language, and how annotator proficiency affects the distribution given the same text.

\paragraph{Collecting Labels}
Taking English as a pivot language, we employ native English speakers (``NT'') and bilingual raters whose mother language is Mandarin Chinese, including 5 professional translators (``PT''), 5 overseas students (``OS''), and 5 general users (``GU''). Each annotator assigns labels to 100 examples sampled from each of \ustsu and \ustsc (the same data set used in \secref{sec:secondcasestudy}), which have  been manually post-edited by professional translators to ensure content alignment.

\paragraph{Results} 
We average the \kl between collective distributions drawn from 19 raters given \textit{zh} text, and 5 native English speakers (NT) given \textit{en} text. 
\tabref{tab:kl-language-prof} shows  
there is not a substantial distributional divergence. Differences decline further as annotations of the other three groups of bilingual raters 
are incorporated.

\begin{table}[t!]
\centering
    \begin{tabular}{l c c c c}
        \toprule
        \textit{en}-rater & NT & +PT & +OS & +GU \\
        \midrule
        \ustsu & 0.69 & 0.67 & 0.53 & 0.38 \\
        \ustsc & 0.94 & 0.78 & 0.73 & 0.68 \\
        \bottomrule
    \end{tabular}
\caption{KL-divergence of labels as ratings from less proficient language speakers are incorporated.}
\label{tab:kl-language-prof}
\end{table}

Detailed analysis of distributions across each of these groups (\figref{fig:multilingualroles}) reveals that the language of the text affects the distribution of human opinions.
On both \ustsu and \ustsc, the distribution differs substantially between native Chinese speakers and native English speakers when given \textit{zh} and \textit{en} sentence pairs, respectively.
While the \textit{zh} annotations cluster in the lower $\sigma$ region, those for \textit{en} are dispersed across a large $\sigma$ span.

\figref{fig:multilingualroles} also shows that the distribution of professional translators mirrors that of English natives, while general users differ substantially from both these groups, but are similar to native-speaker Chinese annotators who are given \textit{zh} text.
We suspect that translators make judgements based on the meaning of \textit{en} text directly, but general users may 
use translation tools to translate \textit{en} text back to \textit{zh} to support their understanding, meaning they are in fact rating a Chinese text pair.
Intermediate-level overseas students may mix strategies and thus are somewhere in between these two extremes.

\textbf{Discussion}
The differences we observe may be attributed to bias introduced during manual translation.
Each sentence in a pair is translated separately, so while a source pair may have lexical overlap, this may not carry over under independent translation. 
We examine this effect by calculating the word overlap similarity as \eqnref{eq:sim} for \textit{zh}/\textit{en} pairs, where $T_1$ and $T_2$ are whitespace-tokenised words for English and based on the \textit{jieba segment tool} for Chinese.
We calculate string similarity as:
\begin{align}
    Sim = \frac{len(T_1\cap T_2) + 1}{max(len(T_1),len(T_2))+1} \label{eq:sim}
\end{align}
As detailed in \tabref{tab:casestudyStringsim}, the lexical overlap similarity for \textit{en} and \textit{zh} is similar for \ustsu and \ustsc, suggesting that inconsistencies under translation are not a primary cause of the observed discrepancy.
\begin{table}[t!]
\centering
    \begin{tabular}{l c c c}
        \toprule
        lan & \ustsu & \ustsc & \usts \\
        \midrule
        zh & 0.42 & 0.45 & 0.44 \\
        en & 0.40 & 0.43 & 0.41 \\
        \bottomrule
    \end{tabular}
\caption{Lexical similarity between \textit{en} and \textit{zh} pairs sampled from \ustsu, \ustsc, and the combination of the two.}
\label{tab:casestudyStringsim}
\end{table}

\paragraph{In summary}
The language of the text impacts the distribution of human judgements.
In our analysis, English results in higher-uncertainty labelling than Chinese, for both uncontroversial and contentious cases.
This suggests that the previous assumption that labels remain identical across languages as long as the meaning of the text is kept the same, is potentially problematic, even though pairwise lexical overlap remains similar.

%% file: section/discussion.tex
\section{Discussion}
We focus on the STS task in this work. However, the methods we propose can be transferred to other subjective textual regression tasks, such as sentiment analysis (SA) rating and machine translation quality estimation in the format of direct assessment (DA).
Similar findings stemming from task subjectivity may be relevant to other types of NLP tasks relying on human annotation. High disagreement among annotators may occur due to ambiguous labelling, where it is challenging to compile guidelines that are widely accepted and consistently interpreted by all individual annotators.

In practice, it may be difficult to estimate the distribution of human annotations in instances where multiple annotators are difficult to source, such as occurs in clinical and biomedical STS due to the need for highly specialised knowledge. 
Transfer learning, which relies on patterns learned from general-purpose USTS, provides a means to predict such a distribution, if noisily.
We propose to explore the direct estimation of individual ratings by in-context learning based on large language models (LLMs), e.g.\ GPT-3~\citep{brown2020gpt3} and ChatGPT.\footnote{\url{https://openai.com/blog/chatgpt}}
LLMs are able to perform in-context learn --- perform a new task via inference alone, by conditioning on a few labelled pairs as part of the input~\cite{min-etal-2022-rethinking}.

ChatGPT appears to be highly effective at style transfer and tailoring of content to specific audiences such as \textit{five-year old children} or \textit{domain experts}, through learning about language style and tone from interactional data and individual preferences.
This allows it to generate more personalised responses~\citep{aljanabi2023chatgpt}.
\citet{deshpande2023toxicity} show that assigning ChatGPT a persona through the parameter \textit{system-role}, such as \textit{a bad/horrible person}, can increase the toxicity of generated outputs up to sixfold.

Additionally, \citet{schick-schutze-2021-generating} show that generative LLMs can be used to automatically generate labelled STS datasets using targeted instructions. This data can be utilised to improve the quality of sentence embeddings. 
Together, these imply that LLMs may have utility in generating personalised semantic similarity assessments, based on annotator meta data (e.g.\ age, educational background, or domain expertise).

Simulating variation in judgements between individual annotators using synthetic personalised ratings could mitigate ineffective training and evaluation caused by aggregation, given that neither a scalar nor a single Gaussian fits the set of observed judgements adequately for instances of high disagreement.

%% file: section/conclusion.tex
\section{Conclusion}
We presented the first uncertainty-aware STS corpus, consisting of 15k Chinese examples with more than 150k annotations. 
The dataset is intended to promote the development of STS systems from
the perspective of capturing inherent disagreements in STS labelling,
and establish less biased and more nuanced gold labels when large variances exist among individual ratings.

We additionally examine the models' ability to capture the averaged opinion and the distribution of collective human judgements. 
Results show that the uncertainty captured by current models is not explained by the semantic uncertainty that results in disagreements among humans. Rather, it tends to reflect the predictive confidence over the whole data set.
We also found that the text language and language proficiency of annotators affect labelling consistency.

%% file: section/acknowledgement.tex
\section*{Acknowledgements}
We thank the anonymous reviewers and editors for their helpful comments, Yanqing Zhao, Samuel Luke Winfield D’Arcy, Yimeng Chen, Minghan Wang in Huawei TSC and NLP Group colleagues in The University of Melbourne for various discussions.
Yuxia Wang is supported by scholarships from The University of Melbourne and China Scholarship Council (CSC).